\documentclass[10pt]{article} 
\usepackage[accepted]{tmlr}
\usepackage{algorithm}
\usepackage{float}
\usepackage{algpseudocode} 

\usepackage{amsmath,amsfonts,bm}

\def\eqref#1{equation~\ref{#1}}

\def\1{\bm{1}}

\DeclareMathAlphabet{\mathsfit}{\encodingdefault}{\sfdefault}{m}{sl}
\SetMathAlphabet{\mathsfit}{bold}{\encodingdefault}{\sfdefault}{bx}{n}

\usepackage{hyperref}
\usepackage{graphicx} 
\usepackage{url}

\title{Clus-UCB: A Near-Optimal Algorithm for Clustered Bandits}

\author{\name Aakash Gore \email aakash.gore@iitb.ac.in \\
      \addr Department of Electrical Engineering\\
      Indian Institute Of Technology Bombay
      \AND
      \name Prasanna Chaporkar \email chaporkar@ee.iitb.ac.in \\
      \addr Department of Electrical Engineering\\
      Indian Institute Of Technology Bombay\\
      }

\begin{document}

\maketitle

\begin{abstract}
We study a stochastic multi-armed bandit setting where arms are partitioned into known clusters, such that the mean rewards of arms within a cluster differ by at most a known threshold. While the clustering structure is known a priori, the arm means are unknown. We derive an asymptotic lower bound on the regret that improves upon the classical bound of \citet{lai1985asymptotically}. We then propose Clus-UCB, an efficient algorithm that closely matches this lower bound asymptotically. Clus-UCB is designed to exploit the clustering structure and introduces a new index to evaluate an arm, which depends on other arms within the cluster. In this way, arms share information among each other. We present simulation results of our algorithm and compare its performance against KL-UCB and other well-known algorithms for bandits with dependent arms. Finally, we address some limitations of this work and conclude by mentioning some possible future research.
\end{abstract}

\section{Introduction}

The multi-armed bandit (MAB) is a foundational problem in probability theory that encapsulates the classic trade-off between \textit{exploration} and \textit{exploitation}. It is typically abstracted as a scenario where a gambler is faced with \( k \) slot machines (arms), each with an unknown reward distribution, and must decide which arm to pull at each timestep to maximize the cumulative reward. The arms are assumed to belong to the same distribution family but with different (and unknown) means.

A seminal contribution in this area is by \citet{lai1985asymptotically}, who showed that any uniformly good algorithm \footnote{A uniformly good algorithm is one which incurs $o(N^a)$ regret for all $a>0$ on all instances}must incur at least \( O(\log N) \) regret, where N is the horizon. Several algorithms such as KL-UCB, UCB, and \( \varepsilon \)-greedy have been proposed that asymptotically attain this logarithmic regret. This framework models arms which are independent of each other.

Bandit problems where arms are correlated or dependent have also been studied in the literature. These fall into the category of structured bandit problems. Many times, information about the structure results in fewer suboptimal arm pulls, and results in lower regret bounds. In this paper, we work with a similar structured bandit problem, specifically one where arms are clustered together.

\subsection{Related Work}

The classical (MAB) problem has received significant attention in the past, with one of the most notable contributions being by ~\citet{lai1985asymptotically}. Using a change-of-measure argument, they derived theoretical lower bounds on the regret incurred by comparing an algorithm's performance on two instances which are similar except the caveat of having different optimal arms. They also proposed a framework for constructing asymptotically efficient algorithms that achieve logarithmic regret.

A closely related work is that of ~\citet{graves1997asymptotically}, where regret bounds were established for bandits in a controlled Markov chain setting. This work generalizes the procedure of finding a lower bound as a linear optimization problem. We use this approach in Section 3 to derive the lower bound for our problem.

For the classical MAB setting with independent arms, several algorithms have been proposed to achieve optimal regret asymptotically. Among the most influential are UCB by ~\citet{auer2002finite}, and KL-UCB by~\citet{garivier2011kl}. KL-UCB works by selecting the arm with the most optimistic estimate of the mean reward, derived from a KL-divergence-based upper confidence bound. Our proposed algorithm is inspired by this principle and extends it to settings that showcase clustering.

Structured bandits, where dependencies among arms are leveraged to minimize regret, have also been explored. For example, ~\citet{combes2014unimodal} and ~\citet{magureanu2014lipschitz} studied bandits under unimodal and Lipschitz structures, respectively, and developed near-optimal algorithms. ~\citet{mersereau2009dynamic} and ~\citet{dani2008stochastic} considered linear bandits, where rewards are assumed to be linear functions of unknown parameters. \citet{zhang2023multiarmed} studied the MAB problem on a graph, where an agent has to maximize the cumulative reward collected from the nodes of a known graph. ~\citet{agrawal1989asymptotically} studied the case of controlled IID processes with a known finite parameter space, and drew parallels between this and a specialized MAB setting. 

~\citet{pandey2007multiarmed} investigated bandits with dependent arms, specifically instances where arms are organized into clusters. They assumed that arm parameters in a cluster are drawn from a known generative model. They formulated a two-level policy assuming that the parameter distribution is tightly centered around its mean. Our problem formulation is a special case of this, where the parameter distribution is uniform over a predefined range.  This makes the distribution spread out, not tightly centered. This motivates the need for a new algorithm.



\subsection{Our Contributions:}

\begin{itemize}
    \item We introduce a framework where arms are organized into \textit{constrained overlapping clusters}, and derive theoretical lower bounds on regret in this structured bandit setting. By constrained, we mean that the arm means within a cluster cannot differ by more than a known threshold.
    \item We propose \textbf{Clus-UCB}, an algorithm that efficiently exploits this structure and asymptotically achieves the regret lower bound (almost).
    \item We provide both theoretical analysis of the algorithm's performance in the Appendix, and simulation results in a later section, that demonstrate the practical effectiveness and  theoretical optimality of our algorithm.
\end{itemize}

\section{Model and Problem Formulation}
 In this section, we first describe the standard stochastic bandit framework, followed by the specific structure of clustered arms that we address in this work.

\subsection{Stochastic Bandit Framework}

At each round $n = 1, 2, \ldots, T$, a learner selects one of $K$ arms and receives a reward sampled from a distribution(unknown). Each arm $k$ is associated with an unknown parameter $\theta_k \in \Theta$ and a known density $f(x; \theta_k)$ with respect to a measure $\nu$. We assume:
\[
\int |x| f(x; \theta) \, d\nu(x) < \infty, \quad \forall \theta \in \Theta.
\]
The expected reward of arm $k$ is given by:
\[
\mu(\theta) = \int x f(x; \theta) \, d\nu(x).
\]

\textbf{Policy:} A sequence $\pi = (\pi_n)$, where $\pi_n \in \{1, \ldots, K\}$, is admissible if $\pi_n$ is $\mathcal{F}_{n-1}$-measurable (i.e., depends only on past actions and rewards).

Let $\mu^* = \max_k \mu_k$ and denote $T_k^\pi(n)$ as the number of times arm $k$ is pulled upto round $n$ under $\pi$.

\textbf{Regret:} Regret under $\pi$ until round $n$ is: 
\[
R^\pi(n, \nu((\theta)) = \sum_{k: \mu_k < \mu^*} (\mu^* - \mu_k) \mathbb{E}[T_k^\pi(n)].
\]
Here, $\theta$ is the parameter vector and $\nu(\theta)$ is the
instance.

\subsubsection{KL Divergence}

For densities parameterized by $\theta$ and $\vartheta$:
\[
I(\theta, \vartheta) = \int \log \left( \frac{f(x; \theta)}{f(x; \vartheta)} \right) f(x; \theta) \, d\nu(x).
\]

The one-sided KL divergence is defined as:
\[
I^+(\theta, \vartheta) = 
\begin{cases}
I(\theta, \vartheta) & \text{if } \mu(\theta) < \mu(\vartheta), \\
0 & \text{otherwise.}
\end{cases}
\]
\textbf{Assumptions:}
\begin{itemize}
    \item $0 < I(\theta, \vartheta) < \infty$ if $\mu(\vartheta) > \mu(\theta)$,
    \item $I(\theta, \vartheta)$ is continuous in $\mu(\vartheta)$.
\end{itemize}

For Bernoulli distributions with means  $\theta$ and $\vartheta$,
\[
I(\theta, \vartheta) = \theta \log \left( \frac{\theta}{\vartheta} \right) + (1 - \theta) \log \left( \frac{1 - \theta}{1 - \vartheta} \right).
\]

\subsubsection{KL-UCB Algorithm}

For each arm $k$, define the KL-UCB for $n^{\rm th}$ round as:
\[
\sup \{\vartheta :T_k(n) \cdot I(\hat{\theta}_k(n), \vartheta) \leq \log n + a\log \log n\},
\]
where $\hat{\theta}_k(n)$ is the empirical mean of $k^{\rm th}$ arm in $n^{\rm th}$ round, and $a$ is a constant greater than 3.
At each round, select the arm with the highest KL-UCB. Note that each arm must be pulled at least once, for the empirical means to be defined.

\subsection{Clustered Arm Structure}

We now introduce the structure in which arms are grouped into overlapping clusters. Let $c \in \{1, \ldots, M\}$ index clusters, and let $K_c$ denote the number of arms in cluster $c$. For the remainder of this work, we analyze a Bernoulli bandit, however, as in the case of the KL-UCB algorithm, we believe this can be extended to the exponential family. For simplicity, we assume only one unique best arm.

The clustering structure is given as:  
For two arms $i, j$ belonging to the same cluster $c$, we have:
\[
|\mu_c^i - \mu_c^j| < \beta_c, \quad \text{for all } i, j \in \{1, \ldots, K_c\}, \quad c \in \{1, \ldots, M\}.
\]
where $\beta_c > 0$ is known for each cluster. The assumption of known cluster widths is not purely of theoretical interest, but is also practical in cases where a rough estimate or a non-trivial upper bound on the width is available.
The following are some important and notational points to keep in mind:

\begin{itemize}
    \item $\mu_c^k$ is the mean reward of arm $k$ in cluster $c$.
    \item $c^*$ is the optimal cluster, i.e., the cluster containing the arm with the highest mean.
    \item Each arm might satisfy the clustering property for multiple clusters. For example, an arm may fall at the intersection of the allowed spaces of two clusters. However, we only know its allegiance to one of these clusters. This is showcased in Figure 1.    
    
\end{itemize}

Define $\Theta$ as the set of all mean vectors in $[0, 1]^K$ that satisfy the above clustering condition.

\begin{figure}
\begin{center}
\includegraphics[scale=0.7]{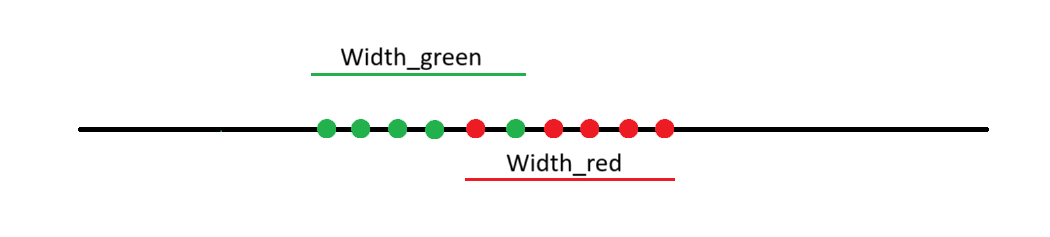}

\end{center}
\caption{The widths represent cluster spans. While the fifth green point also lies within the red span, it's labeled green. Similarly, the first red point also falls within the green span but is labeled red.}

\end{figure}

\section{Lower Bound for Regret}
In this section, we state an asymptotic (when $T$ grows large) regret lower bound satisfied by any good algorithm $\pi \in \Pi$. An algorithm $\pi$ is good if $R^{\pi}(n,\nu(\theta)) = o(n^{\alpha})$ as $n$ grows large for all $\alpha > 0$ and for all $\theta \in \Theta$.

\textbf{Theorem 1:} Let $\pi \in \Pi$ be a uniformly good rule. For any $\theta \in \Theta$, we have:
\[
\liminf_{T \rightarrow \infty }\frac{R^{\pi}(T,\nu(\theta))}{\log T} \geq C(\theta),
\]
where
\[
C(\theta) = \sum_{c=1, c \neq c^*}  \min\left( \sum_{k=1}^{K_c} \frac{\mu^* - \mu^k_{c}}{\alpha_c^k L_c}, \frac{\mu^* -\mu_c^{min}}{I^+(\mu_c^{min}, \mu^*-\beta_c)}\right)  + \sum_{k \neq k^*} \frac{\mu^* - \mu_{c^*}^{k}}{I^+(\mu_{c^*}^{k}, \mu^*)},
\]
\begin{itemize}
    \item $\alpha_c^{k} = I^+(\mu^k_c, \mu^*) - I^+(\mu^k_{c}, \mu^* - \beta_c)$,
    \item $b^k_{c} = I^+(\mu^k_{c}, \mu^* - \beta_c)$,
    \item $L_c = 1 + \sum_{k=1}^{K_c} \frac{b^k_{c}}{\alpha^k_{c}}$,
    \item $\mu_c^{min} = \min_{k}{\mu_c^k}$.
\end{itemize}

The lower bound of regret is derived using the results of controlled Markov chains from \citet{graves1997asymptotically}. The proof is presented in the Appendix. In general, the main idea used for deriving lower bounds is to consider an alternate instance, which is 'close' to the actual instance under consideration, but has a different optimal mean. The agent needs to explore sufficiently to distinguish between the original instance and the alternate instance. For the clustered case under consideration, the alternate instance parameters also belong to the clustered parameter space, unlike the classical case, where the parameters were unconstrained. This fact results in a different lower bound in the structured case. In the proof, we consider an alternate instance as mentioned earlier, and the terms in the lower bound arise naturally as a result of the cluster constraints. The following are some important points: 
\begin{itemize}
\item The exploration term for an arm, i.e., $\alpha_c^k L_c$, is dependent upon the means of other arms in that cluster. This is in contrast to the classical regret bound, where this term only depends on that arm's mean.
    \item This regret bound is always lower than that of the classical bandits derived in \citet{lai1985asymptotically}.
    \item For arms belonging to $c^*$, the regret term is the same as that of classical bandits. This is because inside $c^*$, the cluster structure makes no difference. On the other hand, for suboptimal clusters, we exploit the structure and make improvements in the bound.
    \item It is seen that the regret contains a $\min(a,b)$ term. The second argument in this corresponds to the regret incurred by only pulling the worst arm in a cluster. All other arms in the cluster must be pulled sub-logarithmic times in expectation. Intuitively, the second term corresponds to instances where it is relatively easier to distinguish the minimum arm in a cluster from the best arm in the instance, while the first term corresponds to instances where the agent must pull all arms in the cluster to be certain of its sub-optimality. Hence, it is likely that if we have a loosely constrained cluster, the first term would be the minimum, while for tight clusters, the second term would be the minimum. However, in most of our presented here or otherwise, we found that the first term appears.

    \item Note that for the trivial case of $\beta_c = 1$, we essentially have no clustering information. Thus, the lower bound term for that cluster becomes the same as in the classical case. Here, we abuse notation to convey that 
    \[
   \frac{\mu^* -\mu_c^{min}}{I^+(\mu_c^{min}, \mu^*-\beta_c)} = \frac{\mu^* -\mu_c^{min}}{0} = \infty, 
    \]
    and
    \[
   \sum_{k}\frac{\mu^* - \mu^k_{c}}{\alpha_c^k L_c} = \sum_{k}\frac{\mu^* - \mu^k_{c}}{I({\mu}_c^k, \mu^*) + \sum_{k' \in c, k' \neq k}   I^+({\mu}_{c}^{k'}, \mu^* - \beta_c)} = \sum_{k}\frac{\mu^* - \mu^k_{c}}{I({\mu}_c^k, \mu^*)}.
    \]
    \item For the trivial case of $\beta_c = 0$, we have $K_c$ arms with the same mean. Hence, $\mu_c^{min} = \mu_c^k = \mu $(let). This essentially means that we have only one arm in the cluster.
    \[
   \frac{\mu^* -\mu_c^{min}}{I^+(\mu_c^{min}, \mu^*-\beta_c)} = \frac{\mu^* -\mu}{I^(\mu, \mu^*)}, 
    \]
    and
    \[
   \sum_{k}\frac{\mu^* - \mu^k_{c}}{\alpha_c^k L_c} = \sum_{k}\frac{\mu^* - \mu}{I({\mu}, \mu^*) + \sum_{k' \in c, k' \neq k}   I({\mu}, \mu^*)} = \frac{\mu^* - \mu}{I({\mu}, \mu^*)}.
    \]   
\end{itemize}

\section{Clus-UCB Algorithm }
\noindent\vspace{0pt}
We now present an algorithm whose regret closely matches the lower bound derived in the previous section for clustered overlapping bandits.

\begin{algorithm}
\caption{Clus-UCB Algorithm}
\begin{algorithmic}
\State \textbf{Input:} Total time steps $T$, number of clusters $M$, number of arms $K_c$ in each cluster, total number of arms $K$, arm cluster pairs, a constant $a \geq 5$
\State Pull each arm once to initialize 
\For{$n = K+1$ to $T$}
    \State Compute the empirical mean $\hat{\mu}_{k}^{c}(n)$ for each arm $k$ in cluster $c$
    \State Let $t_c^k(n)$ denote the number of times arm k in cluster c has been pulled up to timestep n
    \For{each arm $k$ in clusters $c \neq c_{\max}$}
        \State Compute the Clus-UCBs:
        \[
        v_{k}^{c}(n) =  \sup\left \{ q : t_c^k(n) I^(\hat{\mu}_c^k(n), q) + \sum_{k' \in c, k' \neq k} t_{c}^{k'}(n)  I^+(\hat{\mu}_{c}^{k'}(n), q - \beta_c) \leq \log n + a\log \log n \right\}
        \]
        
    \EndFor
    \State Select arm $k_n = \arg\max_{k,c} v_{k}^{c}(n) $
    \State Pull arm $k_n$ and observe reward
\EndFor
\end{algorithmic}
\end{algorithm}
\textbf{Theorem 2:} Assuming that the bandit arms are Bernoulli and  clustered according to Section 2.2, Clus-UCB's asymptotic regret is upper bounded as

\[
\liminf_{T \rightarrow \infty }\frac{R^{\pi}(T,\nu(\theta))}{\log T} \leq C(\theta),
\]
where
\[
C(\theta) = \sum_{c=1, c \neq c^*}   \sum_{k=1}^{K_c} \frac{\mu^* - \mu^k_{c}}{\alpha_c^k L_c}  + \sum_{k=1, k \neq k^*} \frac{\mu^* - \mu_{c^*}^{k}}{I^+(\mu_{c^*}^{k}, \mu^*)}.
\]
The proof of Theorem 2 is present in the appendix. The following are some key points about this algorithm:

\begin{itemize}
    \item \textbf{Rare Suboptimal Pulls Due to Confidence Underestimation:} 
    The event in which an arm of a suboptimal cluster is pulled because the Clus-UCB of the optimal arm falls below its mean occurs only $O(\log \log T)$ times.
    
    \item \textbf{Pull Ratio Among Arms in a Suboptimal Cluster:} 
    Within a suboptimal cluster, arms are pulled in inverse proportion to their exploration coefficients . That is, for two arms with exploration parameters $\alpha^c_{k_1}$ and $\alpha^c_{k_2}$, the expected number of times they are pulled over a long time satisfies:
    \[
    E[t_c^{k_1}] : E[t_c^{k_2}] \approx \alpha_c^{k_2} : \alpha_c^{k_1}
    \]
    \item \textbf{Expected Pulls of Arms in Suboptimal Clusters:} 
    An arm $k$ belonging to a suboptimal cluster $c$ is pulled approximately 
    \[
    \frac{\log T}{\alpha_c^{k} \cdot L_c} + O(\log \log T).
    \]
    times in expectation, over a long time, where $\alpha_c^{k}$ and $L_c$ are as defined earlier.

    \item \textbf{Near-Optimality:} 
    The upper bound presented above, matches the regret lower bound derived earlier on most instances, but not all. This makes the algorithm near-optimal. 
\end{itemize}
The motivation to use the Clus-UCB index as done in the algorithm is through the lower bound derived and the analysis done by \cite{garivier2011kl}. In the appendix, a similar approach is taken to prove Theorem 2, along with a few modifications.
\section{Simulation Results and Discussion:}
We ran simulations for comparing KL-UCB, Clus-UCB and a KL-UCB-based Two-level-Policy(TLP) on different bandit instances. Figures 2-5 show the results. All experiments were performed on a computer with 16 gigabytes of RAM. No GPU was used. The plots shown are the average of 48 simulations. To speed up the simulations, we used a multiprocessing framework with 16 CPU cores. Furthermore, we updated the UCBs every 50 timesteps to reduce computation time. The UCBs were calculated using binary search, and are accurate up to 4 decimal places. Here, $\beta$ is the cluster width vector.

We consider two variants of the Two-Level Policy (TLP) suggested by \citet{pandey2007multiarmed}: MEAN and MAX.
TLP treats each cluster as a “super arm” and uses a base policy (KL-UCB) to choose which cluster to play. Once a cluster is selected, the base policy is applied to its arms. Cluster selection requires a reward estimate:
\begin{itemize}
    \item In MEAN, this is the total successes of all arms in the cluster divided by the total cluster pulls.
    \item 
In MAX, it is the maximum empirical mean among the cluster’s arms.

\end{itemize}
In our experiments, Clus-UCB consistently outperforms KL-UCB. However, on certain instances, TLP can outperform Clus-UCB. That said, TLP is not asymptotically optimal, as it lacks knowledge of cluster widths. Moreover, since TLP assumes arm parameters are tightly clustered, it is straightforward to construct hard instances where its performance degrades sharply (see Figure 5).
This is because the algorithm is unable to distinguish a high mean, low variance, suboptimal cluster from a low mean, high variance optimal cluster quickly.

\section{Misspecification of Cluster Widths}
An important point to consider is the misspecification of cluster widths. Cases where the exact widths are not known, but an estimate is available, might be more practical.
If the widths are overestimated, the proposed algorithm continues to outperform KL-UCB.
The case of underestimated widths, however, is more nuanced.
In the proof of Clus-UCB’s optimality (Appendix), we divide the total number of pulls of an arm in a suboptimal cluster into two cases:
\begin{itemize}
    \item when the Clus-UCB index of the optimal arm is less than its mean, and
    \item when the index is greater than or equal to its mean.
\end{itemize}

We bound these two terms separately.
It is noteworthy that the cluster constraint is used only in bounding the first term, which leads to an $O(\log \log T)$ bound.
Moreover, this bound depends solely on the width of the optimal cluster.
In fact, throughout the proof, there is no requirement that the other (suboptimal) clusters satisfy their respective constraints.

At first glance, this may appear surprising.
However, the problem formulation we present is actually a special case of a more general setting of the allowed instances:
every cluster has an associated width, but the cluster constraint is required to hold only for the optimal cluster in the fixed instance, while suboptimal clusters may violate it.
Our formulation imposes the stricter condition that all clusters satisfy their constraints, which is a subset of the general case.
This is distinct from the scenario where, in a given instance, a fixed cluster (which happens to be optimal) satisfies the constraint but an originally suboptimal cluster which becomes optimal in an alternate instance need not satisfy this constraint.
In the general setting, while constructing an alternate instance, the originally suboptimal cluster may become optimal and must then satisfy its width constraint.
In our setting, the regret lower bound derived earlier continues to apply to this more general case as well.

From this perspective, underestimating the width of a suboptimal cluster does not harm performance—in fact, it can improve the regret bound due to the larger denominator ($\hat{\beta}_c < \beta_c$). This is shown in Figures 6-7, where the cluster width estimate for the suboptimal cluster is reduced from overestimated to underestimated.
However, underestimating the width of the optimal cluster can lead to substantial regret, possibly linear. This is because the $O(\log \log T)$ bound may not hold now. This is shown in Figure 8.
If all cluster widths are specified correctly or are overestimated, Clus-UCB retains the property of being uniformly good. Hence, for the algorithm to work well, the sufficient condition is that the optimal cluster must have an overestimated width.

\begin{figure}
    \centering
    \includegraphics[scale=0.37]{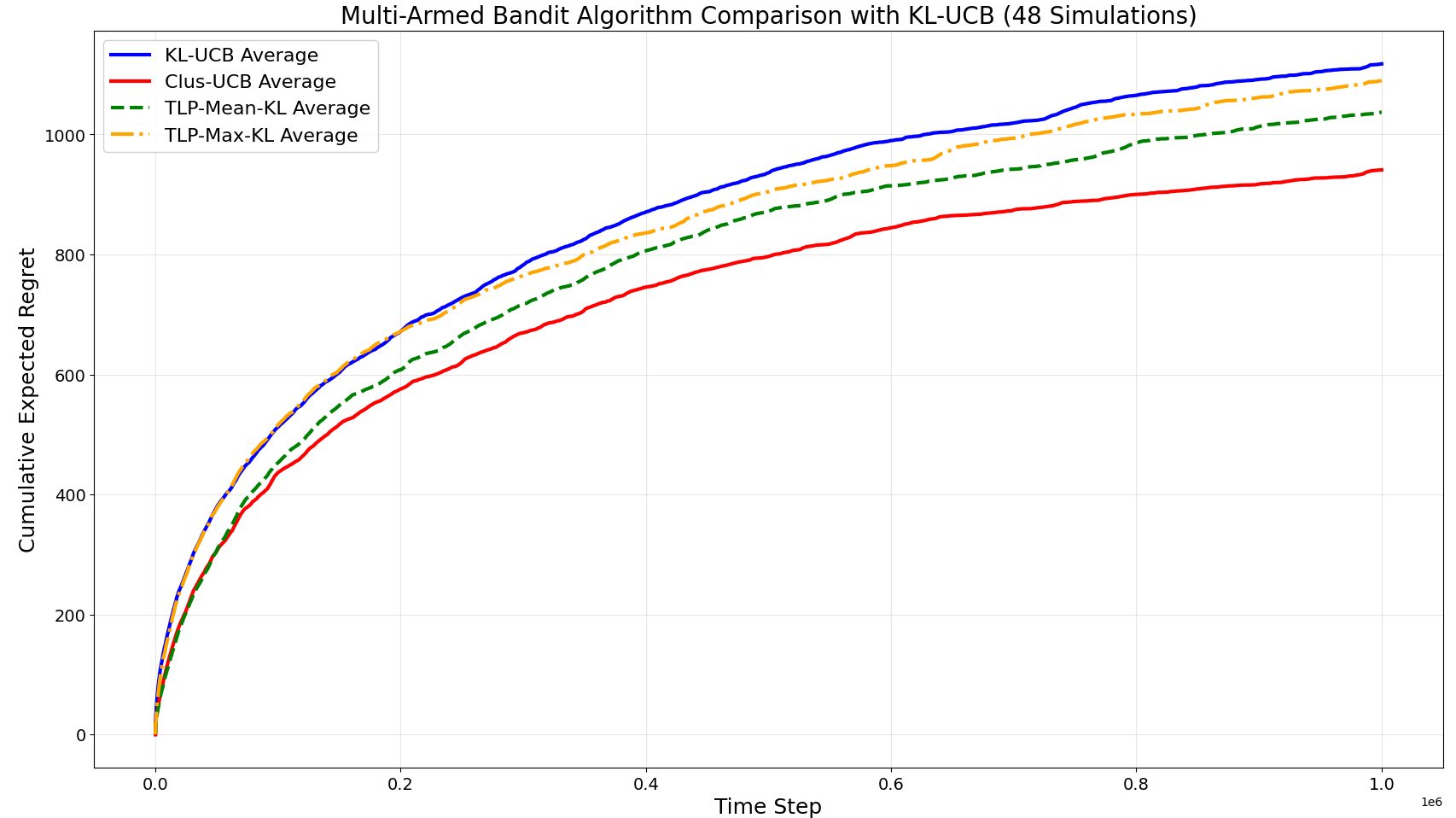}
    \caption{Comparison of Clus-UCB and KL-UCB on the instance [0.40,0.41,0.42], [0.60,0.61,0.62] with $\beta= [0.02,0.02]$ and a horizon of $10^6$ time steps. These represent well separated clusters. The first term in $\min$ appears in the regret lower bound for the suboptimal cluster in this instance.}
    \label{fig:enter-label}
\end{figure}

\begin{figure}
    \centering
    \includegraphics[scale=0.37]{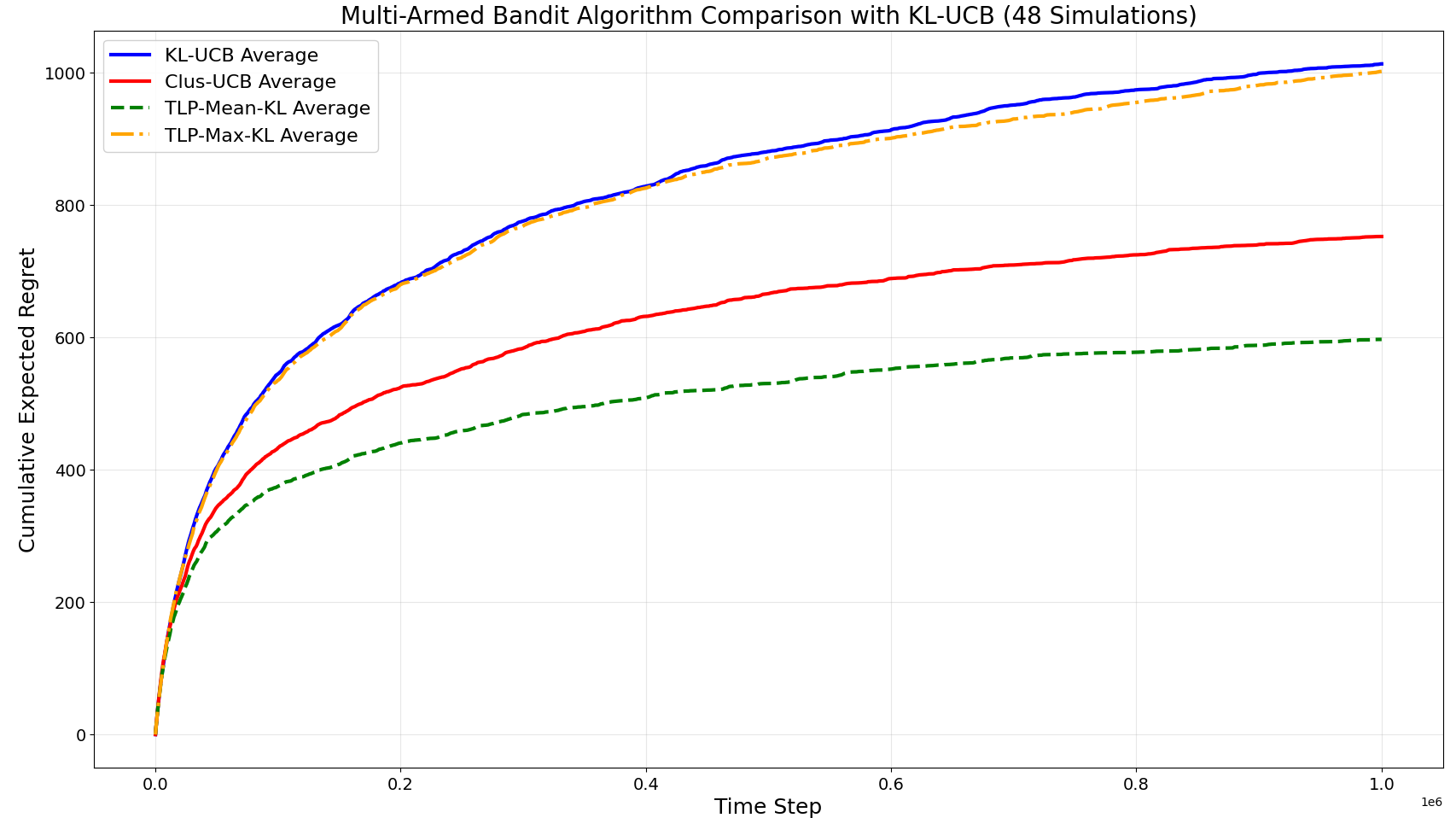}
    \caption{Comparison of Clus-UCB and KL-UCB on the instance [0.80,0.82,0.84], [0.81,0.83,0.85] with $\beta= [0.02,0.02]$ and a horizon of $10^6$ time steps. These represent overlapping clusters. The first term in $\min$  appears in the regret lower bound for the suboptimal cluster in this instance.}
    \label{fig:enter-label}
\end{figure}

\begin{figure}
    \centering
    \includegraphics[scale=0.37]{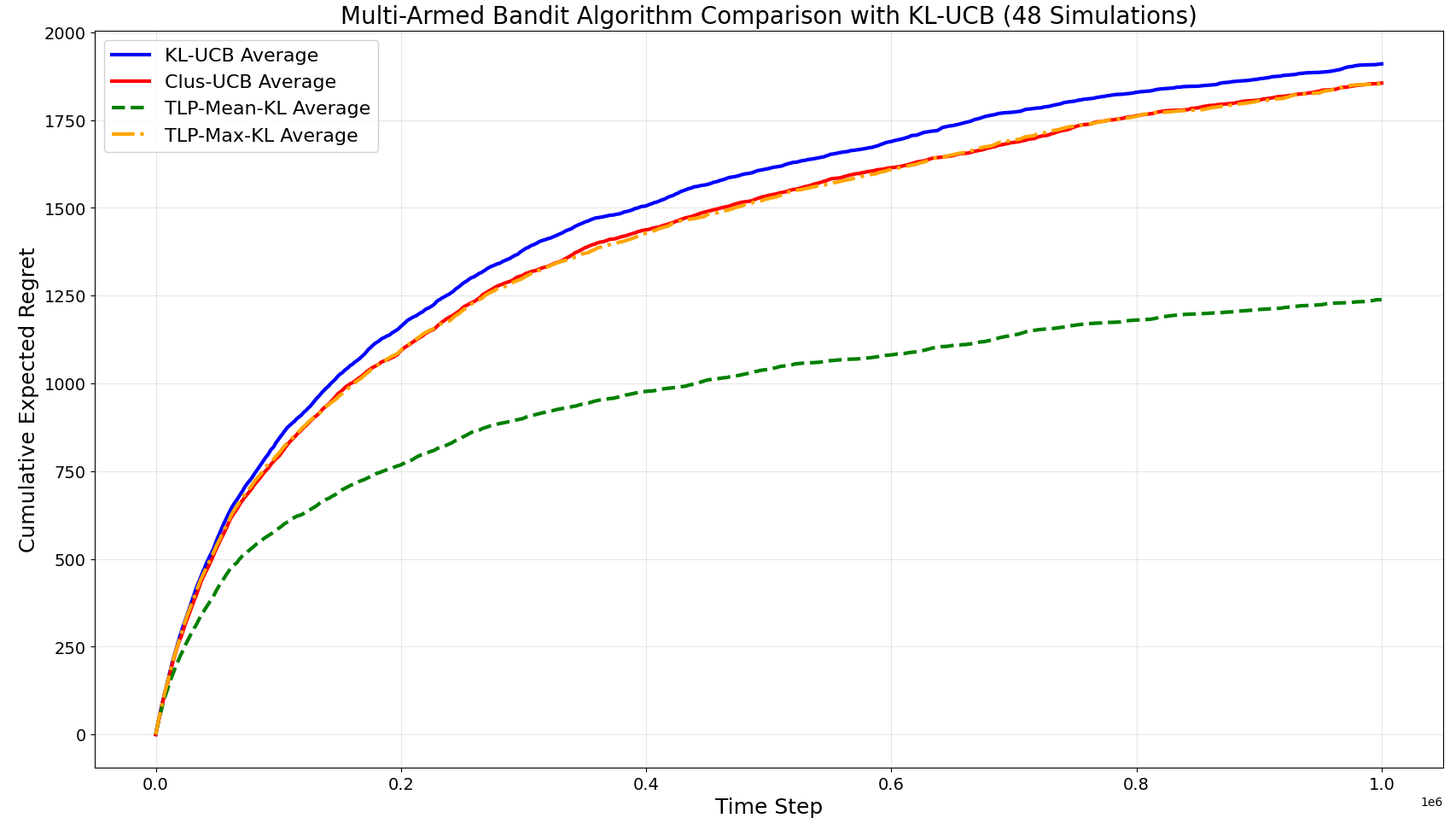}
    \caption{Comparison of Clus-UCB and KL-UCB on the instance [0.41,0.42,0.43], [0.43,0.44,0.45] with $\beta= [0.03,0.04]$ and a horizon of $10^6$ time steps. These represent close but separated clusters. The first term in $\min$ appears in the regret lower bound for the suboptimal cluster in this instance.}
    \label{fig:enter-label}
\end{figure}

\begin{figure}
    \centering
    \includegraphics[scale=0.37]{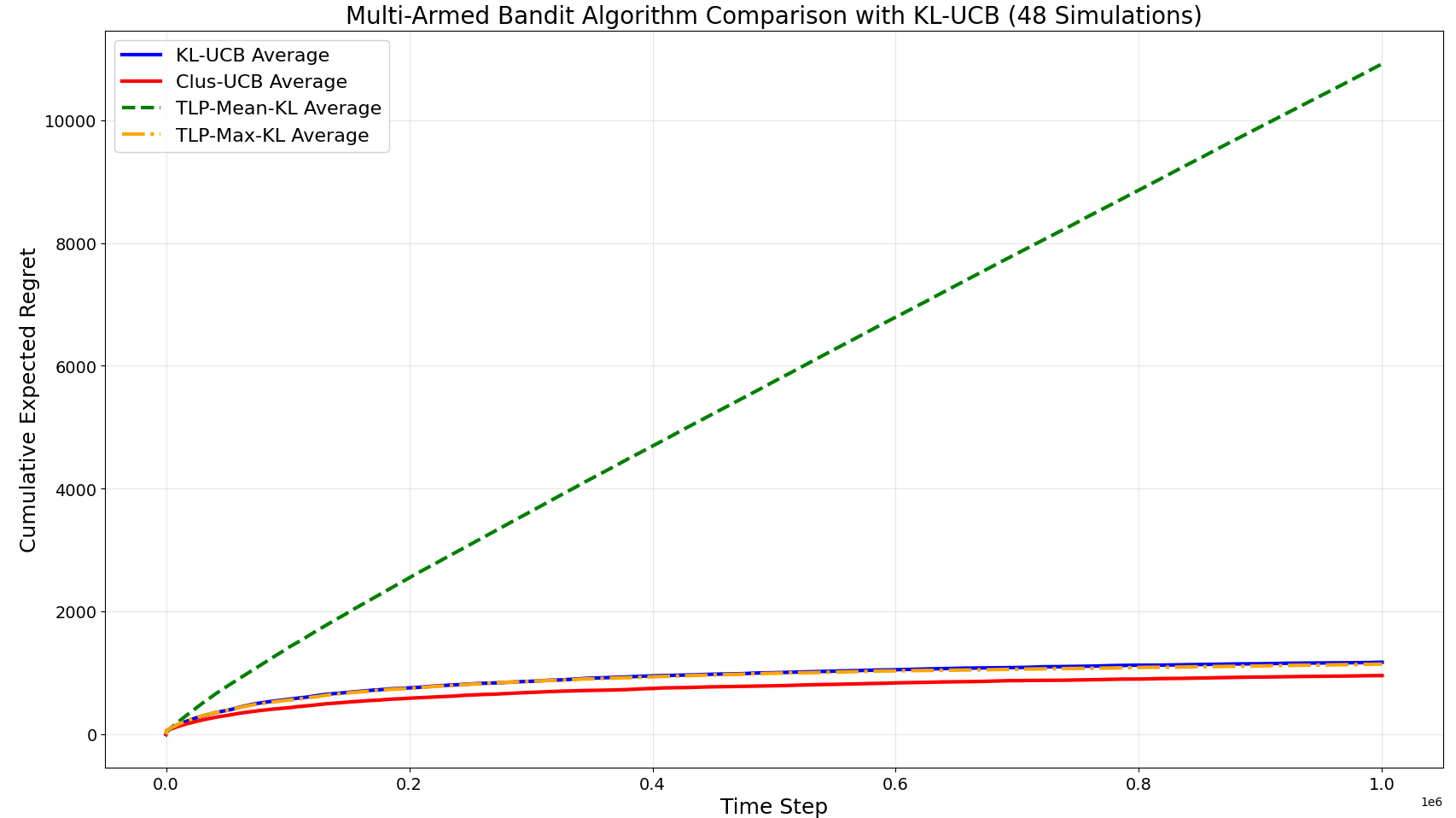}
    \caption{Comparison of Clus-UCB and KL-UCB on the instance [0.68,0.69,0.67], [0.1,0.2,0.7] with $\beta= [0.02, 0.8]$ and a horizon of $10^6$ time steps. This represents an instance where the TLP-Mean policy performs poorly. The first term in $\min$ appears in the regret lower bound for the suboptimal cluster in this instance.}
    \label{fig:enter-label}
\end{figure}

\begin{figure}
    \centering
    \includegraphics[scale=0.37]{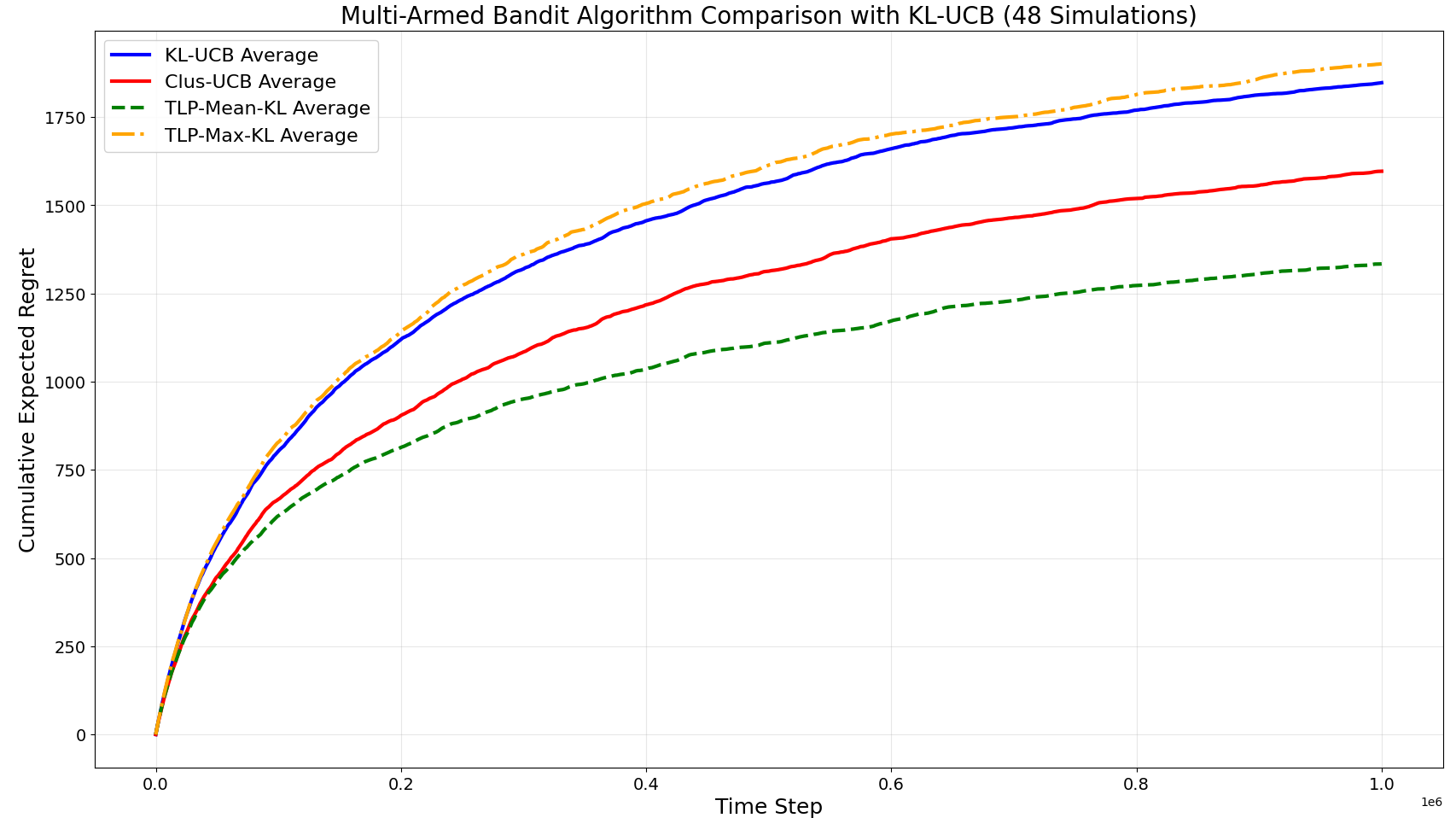}
    \caption{Comparison of Clus-UCB and KL-UCB on the instance [0.41,0.42,0.43], [0.43,0.44,0.45] with $\beta= [0.02,0.02]$ and a horizon of $10^6$ time steps. These represent close but separated clusters. The second term in $\min$ appears in the regret lower bound for the suboptimal cluster in this instance.}
    \label{fig:enter-label}
\end{figure}

\begin{figure}
    \centering
    \includegraphics[scale=0.37]{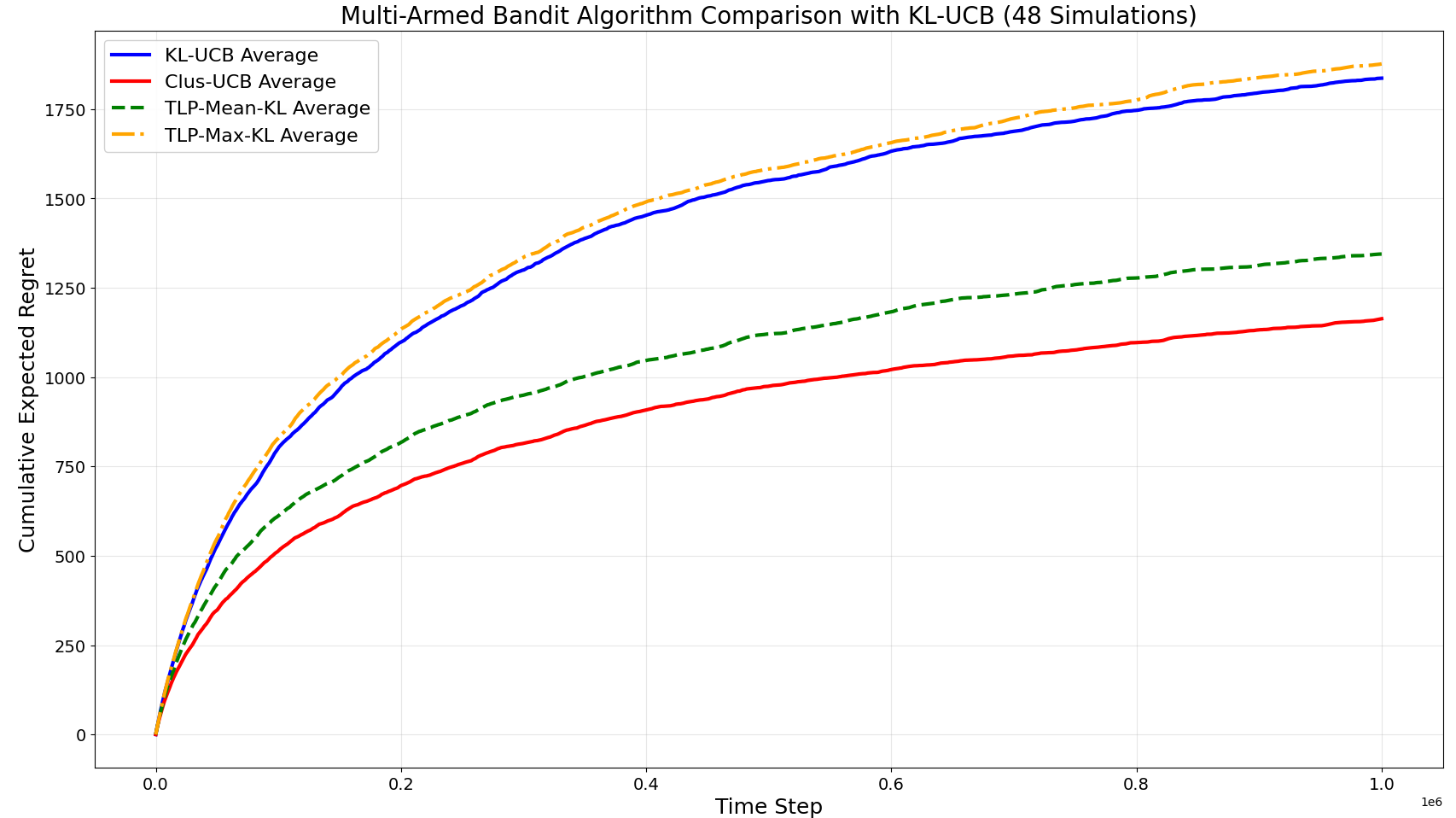}
    \caption{Comparison of Clus-UCB and KL-UCB on the instance [0.41,0.42,0.43], [0.43,0.44,0.45] with $\beta= [0.00,0.02]$ and a horizon of $10^6$ time steps. These represent close but separated clusters. The second term in $\min$ appears in the regret lower bound for the suboptimal cluster in this instance.}
    \label{fig:enter-label}
\end{figure}

\begin{figure}
    \centering
    \includegraphics[scale=0.37]{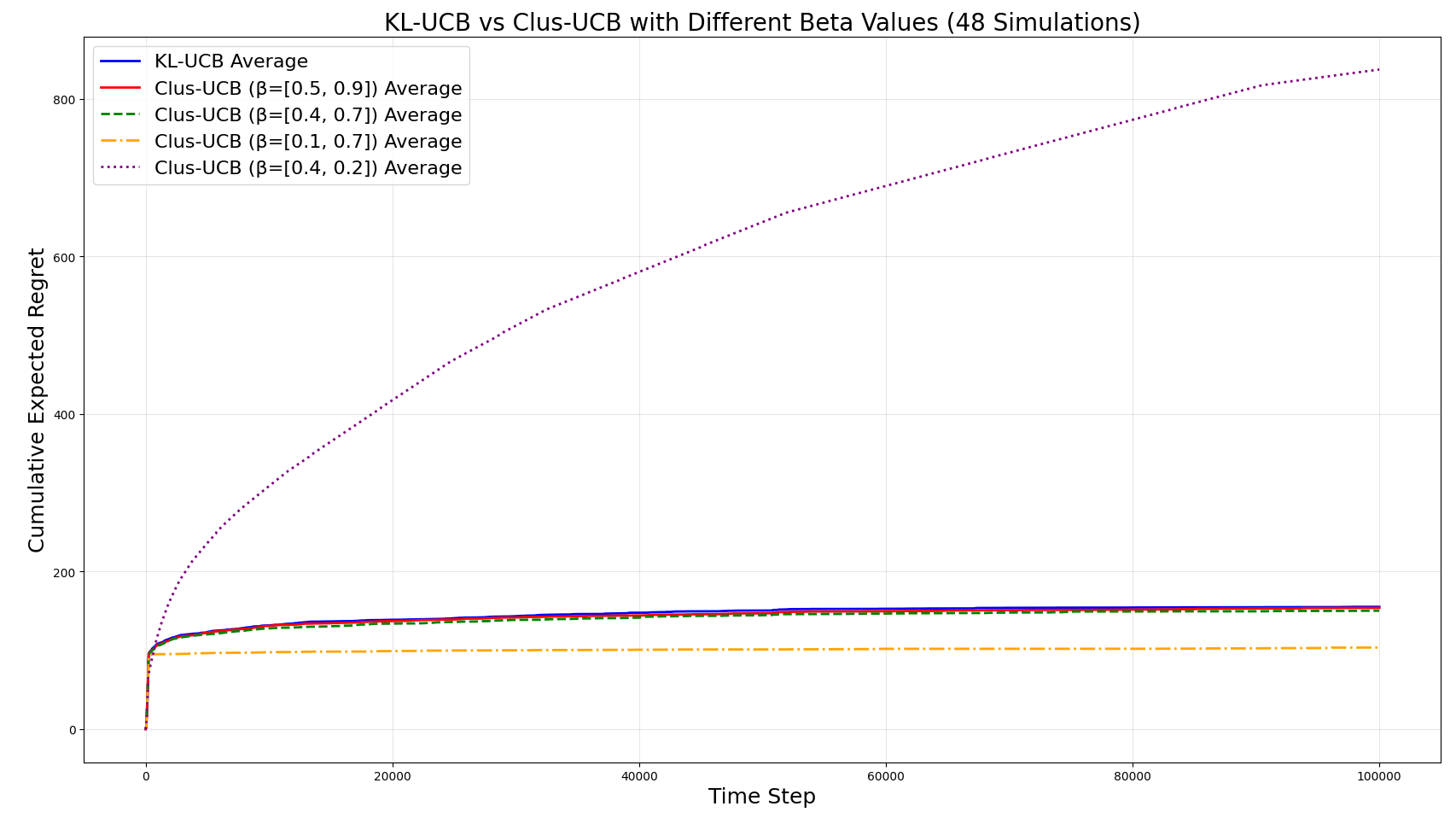}
    \caption{Comparison of Clus-UCB and KL-UCB on the instance [0.3,0.7],[0.1,0.2,0.8] with $\beta=[0.5, 0.9]$, $\beta=  [0.4, 0.7]$, $\beta= [0.1, 0.7]$, and $\beta= [0.4, 0.2] $  and a horizon of $10^6$ time steps.}
    \label{fig:enter-label}
\end{figure}

\section{Limitations and Future Work:}
 Even though the only requirement is the availability of an upper bound on the optimal cluster's width, there might be cases where this isn't available. Thus, this analysis would not apply to such cases, and a different notion of clustering is needed. Note, it is not possible to 'learn' cluster widths to achieve asymptotic optimality, because cluster widths can be arbitrary. Any classical bandit problem could then be framed as a clustered bandit problem, and the regret any algorithm incurs must at least match the bound given by \citet{lai1985asymptotically}.
 
 Furthermore, the algorithm provided is asymptotically optimal on most instances, but not all. However, we believe that a more carefully chosen optimistic index, might perform optimally on all instances, albeit with increased complexity of analysis. It is also possible to develop a randomized Bayesian algorithm, similar to Thompson sampling. The beliefs would still be Beta distributed, but only supporting parameter values which satisfy the clustering constraint.  We leave the proof of optimality of this algorithm as future work. Finally, we have proved the results for Bernoulli bandits, however, a more general analysis for the exponential family, as done for KL-UCB, is applicable here as well.

\section{Conclusion:}
In this work, we derived an improved regret lower bound as compared to the one given by \citet{lai1985asymptotically} for bandits that showcase arm clustering. We assumed the structure of constrained clusters and proposed the Clus-UCB algorithm
to exploit this dependency in arms. We have also shown the near-optimality of the proposed algorithm and run simulations showcasing its advantages over structure-unaware algorithms. We also compare it with the two-level-policy suggested by \citet{pandey2007multiarmed}. We then discuss the cases where the cluster widths are misspecified, and point out a necessary condition for Clus-UCB to perform robustly in misspecified settings. Finally, we discuss some limitations regarding the near-optimality of the algorithm and the assumption of known cluster widths, and discuss the prospects of future work on this. The proofs of all theorems can be found in the Appendix.

\bibliography{main}
\bibliographystyle{tmlr}

\appendix
\section{Appendix}
\subsection{Proof of Theorem 1}
We follow the analytical framework developed by \cite{graves1997asymptotically}. Let $\Theta$ denote the set of all problem instances consistent with the given cluster structure. For each arm $j$, define $\Theta_j$ as the set of instances in which arm $j$ is optimal. Given an instance $\theta \in \Theta$, let $J(\theta)$ be the set of optimal arms under $\theta$.

We define the set of \emph{bad instances} as:
\[
B(\theta) = \left\{ \lambda \in \Theta : \mu_\theta^j = \mu_\lambda^j \ \forall j \in J(\theta), \ \text{and } \lambda \notin \bigcup_{j \in J(\theta)} \Theta_j \right\}.
\]
Here $\mu_{\theta}^j$ is the mean of arm $j$ in the instance $\nu(\theta)$.
Let $C(\theta)$ be the value of the following optimization problem:
\[
C(\theta) = \inf \left\{ \sum_{j \notin J(\theta)} C_j(\mu^*_\theta - \mu_\theta^j) : C_j \geq 0, \ \inf_{\lambda \in B(\theta)} \sum_{j \notin J(\theta)} C_j I(\mu_\theta^j, \mu_\lambda^j) \geq 1 \right\},
\]
where $I(\cdot, \cdot)$ is the KL divergence.

According to Theorem 1 of Graves and Lai, this quantity characterizes the asymptotic lower bound on regret for any uniformly good algorithm $\pi$:
\[
\liminf_{n \rightarrow \infty} \frac{R^{\pi}(n,\nu(\theta))}{\log n} = C(\theta).
\]

Computing $C(\theta)$ reduces to solving a linear program. Suppose that under a bad instance $\lambda$, some arm $i$ from a suboptimal cluster $c_0$ becomes optimal. The value of 
\[
\sum_{j \notin J(\theta)} C_j I(\mu_\theta^j, \mu_\lambda^j).
\]
is minimized when, for all clusters except $c_0$, the arm means under $\lambda$ match those of the suboptimal arms in $\theta$. For cluster $c_0$, the $i$-th arm has a mean greater than $\mu^*_\theta$, while other arms in $c_0$ have means 
\[
\mu_\lambda^j = \max(\mu_\theta^j, \mu^*_\lambda - \beta_c).
\]
The minimum is achieved when $\mu^*_\lambda = \mu^*_\theta$.

For a given cluster $c_0$ with $K_{c_0}$ arms indexed by $k = 1, \ldots, K_{c_0}$, the system of inequalities becomes:
\[
C_i I(\mu_i, \mu^*) + \sum_{k \neq i} C_kI^+(\mu_k, \mu^* - \beta_c) \geq 1,
\]
where $\mu_i = \mu_\theta^i$, and $\mu^* = \mu^*_\theta$.

Let $\mathbf{B}$ be a $K_{c_0} \times K_{c_0}$ matrix, where each column i has elements $b_c^i$. Let $\boldsymbol{\alpha}$ be a diagonal matrix of $\alpha_{c_0}^k$ terms, and $\mathbf{c}$ a column vector of the $C_i$ variables. Define the reward gap vector $\mathbf{a}$ with entries $a_i = \mu^* - \mu_i$.

The linear program becomes:
\[
\begin{aligned}
&\min_{\mathbf{c} \geq 0} \quad \mathbf{a}^\top \mathbf{c} \\
&\text{subject to} \quad (\mathbf{B} + \boldsymbol{\alpha}) \mathbf{c} \geq \mathbf{1}.
\end{aligned}
\]

This optimization can be solved using standard techniques, and we get the desired lower bound.

This process is repeated across all clusters to compute the global infimum $C(\theta)$.

\subsection{Proof of Theorem 2}
This work follows the outline of the proof of Theorem 2 in \citet{garivier2011kl}.
We now state theorem from \citet{magureanu2014lipschitz}.\\
\textbf{Theorem 3:} \\
For all $\delta > k+1$, $n \in \mathbb{N}$, we have:
\[
P\left( \sum_{k=1}^{K} t_c^k(n) I^+\left( \hat{\mu}_c^k, \mu_c^k\right) \geq \delta \right) 
\leq 
e^{-\delta} \left( (\frac{\lceil\delta \log n ]\delta}{k} )^k e^{k+1} \right).
\]
If $\delta = \log n + a \log \log n$ ,with $a \geq 5$, then
\[
\mathbb{E}\left[ \sum_{n=1}^T \mathbb{I}\left\{ \sum_{i=1}^K t_c^k(n) d^+\left(\hat{\mu}_c^k, \mu_c^k\right) \geq \delta \right\} \right] = O(\log \log T)
\]

In the following proof, we haven't explicitly mentioned the dependence of empirical means$(\hat{\mu}(n))$ and number of pulls($t_c^k(n)$ on the time step n. This is done only for neatness. Let 'i' be the best arm in cluster 'c'.
Now, we proceed by bounding the number of pulls as:
\begin{align*}
\mathbb{E}[t_c^i(T)] &= \sum_{n=1}^T \mathbb{I}\{A_n = (i, c)\} \\
&= \sum_{n=1}^T [\mathbb{I}\left\{A_n = (i, c), v^*(n) \geq \mu^*\right\} + \mathbb{I}\left\{A_n = (i, c), v^*(n) < \mu^*\right\}],
\end{align*}

where $v^*(n)$ is the Clus-UCB of the optimal arm.
Now:
\[
\sum_{n=1}^T \mathbb{I}\left\{A_n = (i, c), v^*(n) < \mu^*\right\} \leq \sum_{n=1}^T \mathbb{I}\left\{v^*(n) < \mu^*\right\}
\]
\[
= O(\log \log T) \text{ as proved ahead.}
\]
Note that
\[
\begin{aligned}
t_{c^*}^{k^*} d\left(\hat{\mu}_{c^*}^{k^*}, \mu^*\right) + \sum_{k \neq k^*} t_{c^*}^k d^+\left(\hat{\mu}_{c^*}^k, {\mu}_{c^*}^k\right) 
\geq t_{c^*}^{k^*} d^+\left(\hat{\mu}_{c^*}^{k^*}, \mu^*\right) + \sum_{k \neq k^*} t_{c^*}^k d^+\left(\hat{\mu}_{c^*}^k, \mu^* - \beta_{c^*}\right)
\end{aligned}
\]
as $u_{c^*}^{k} \geq u^* - \beta_{c^*} \quad \forall k \in c^*$. 

Let the left hand side term be A and the right hand side term be B. Therefore,
$A \geq B$

Now $P(B > \delta) \leq P(A > \delta) = O(\log \log T)$ by using Theorem 3.

Therefore, $P(B > \delta) = O(\log \log T)$

\textbf{Other term:}
\[
\sum_{n=1}^T \mathbb{I}\left\{A_n = (i, c), v^*(n) \geq u^*\right\} 
\]
Also note that 
$$
\left\{A_n = (i, c), v^*(n) \geq u^*\right\}
\Rightarrow v_c^i(n) \geq v^*(n) \geq u^*$$
Thus, the inequality continues as
\[
\leq \sum_{n=1}^T [\mathbb{I}\left\{ t_c^i d\left(\hat{\mu}_c^i, \mu^*\right) + \sum_{k \neq i} t_c^k d^+\left(\hat{\mu}_c^k, \mu^* - \beta_c\right) \leq \log n + a \log \log n \right\}\times X_n],
\]
where 
\[
X_n = \mathbb{I}\left\{A_n = (i,c), v_c^i(n) \geq u^* \right\}
\]
\[
\leq \sum_{n=1}^T [\mathbb{I}\left\{t_c^i[ d\left(\hat{\mu}_c^i, \mu^*\right) + \sum_{k \neq i} \frac{t_c^k}{t_c^i } d^+\left(\hat{\mu}_c^k, \mu^* - \beta_c\right)] \leq \log n + a \log \log n \right\}\times X_n]
\]
We now make 2 key observations about the behavior of the algorithm:

\begin{enumerate}
    \item \textbf{Regret upper bound:}  
    The regret of the algorithm can be upper bounded by that of the KL-UCB algorithm. This follows as:
    \[
 \sum_{n=1}^T \mathbb{I}\left\{t_c^i[ d\left(\hat{\mu}_c^i, \mu^*\right) + \sum_{k \neq i} \frac{t_c^k}{t_c^i } d^+\left(\hat{\mu}_c^k, \mu^* - \beta_c\right)] \leq \log n + a \log \log n \right\}\]
\[\leq  \sum_{n=1}^T \mathbb{I}\left\{t_c^i d\left(\hat{\mu}_c^i, \mu^*\right)  \leq \log n + a \log \log n \right\}.
\]

The right hand term is what we get while analyzing KL-UCB. Thus, the regret of Clus-UCB is upper bounded by the regret of KL-UCB, though we will show that it is not tight. Therefore, each suboptimal arm is pulled at most \( O(\log T) \) times.

    \item \textbf{Convergence of Clus-UCB values:}  
    The Clus-UCB values of all suboptimal arms must converge to \( \mu^* \), the mean of the optimal arm. We prove this by contradiction:

    \begin{enumerate}
        \item Suppose the Clus-UCB of a suboptimal arm \( i \) converges to some value \( u < \mu^* \). Then, eventually, the algorithm will stop selecting this arm. As a result, the number of times it is pulled will be sub-logarithmic, contradicting the earlier claim that every suboptimal arm is pulled \( O(\log T) \) times.
        
        \item Suppose instead that the Clus-UCB of a suboptimal arm converges to \( u > \mu^* \). Since the Clus-UCB of the optimal arm converges to \( \mu^* \), the suboptimal arm will eventually have a strictly higher UCB. This would lead the algorithm to pull it linearly often, resulting in linear regret, which contradicts the \( O(\log T) \) upper bound.
    \end{enumerate}
    
\end{enumerate}

Now, notice that
\[
 t_c^i d\left(\hat{\mu}_c^i, v_c^i\right) + \sum_{k \neq i} t_c^k d^+\left(\hat{\mu}_c^k, v_c^i - \beta_c\right) = \log n + a \log \log n, 
\]

\[
 \Rightarrow t_c^i (d\left(\hat{\mu}_c^i, v_c^i\right)-d^+\left(\hat{\mu}_c^i, v_c^i - \beta_c\right)) + \sum_{} t_c^k d^+\left(\hat{\mu}_c^k, v_c^i - \beta_c\right) = \log n + a \log \log n. 
\]
Let $f_i(n) = t_c^i (d\left(\hat{\mu}_c^i, v_c^i\right)-d^+\left(\hat{\mu}_c^i, v_c^i - \beta_c\right))$ and $g_i(n) = \sum_{} t_c^k d^+\left(\hat{\mu}_c^k, v_c^i - \beta_c\right)$

Thus, $ f_k(n) + g_k(n) = \log n +a\log \log n $ for all arms k in cluster c. Also, since we are interested in the time instances when arm 'i' has the maximum cluster index, $g_i(n)\geq g_k(n) \forall k \in 1,2...K_c$
This implies, $f_i(n)\leq f_k(n) \forall k \in 1,2...K_c$
Thus,
\[
\frac{t_c^k}{t_c^i} \geq \frac{d(\hat{\mu}_c^i , v_c^i)- d^+(\hat{\mu}_c^i , v_c^i-\beta_c) }{d(\hat{\mu}_c^k , v_c^k)- d^+(\hat{\mu}_c^k , v_c^k-\beta_c)}.
\]
By the strong law of large numbers, we know that the empirical mean of an arm differs from its true mean by more than $\epsilon$ only finitely many times. Also, $v_c^i \geq \mu*$ and $v_c^i \geq v_c^k$.
Thus,
\[
 \frac{d(\hat{\mu}_c^i , v_c^i)- d^+(\hat{\mu}_c^i , v_c^i-\beta_c) }{d(\hat{\mu}_c^k , v_c^k)- d^+(\hat{\mu}_c^k , v_c^k-\beta_c)} \geq  \frac{d(\mu_c^i-\epsilon , v_c^i)- d^+(\mu_c^i-\epsilon , v_c^i-\beta_c)}{d(\mu_c^k+\epsilon , v_c^i)- d^+(\mu_c^k+\epsilon , v_c^i-\beta_c)}
\]
We also have $\mu_c^i \geq \mu_c^k$, and hence the right hand side is increasing with respect to $v_c^i$
\[
 \Rightarrow \frac{d(\mu_c^i-\epsilon , v_c^i)- d^+(\mu_c^i-\epsilon , v_c^i-\beta_c)}{d(\mu_c^k+\epsilon , v_c^i)- d^+(\mu_c^k+\epsilon , v_c^i-\beta_c)}\geq \frac{\alpha_c^i}{\alpha_c^k}-O(\epsilon).
\]
Thus, 
\[
\sum_{n=1}^T [\mathbb{I}\left\{t_c^i[ d\left(\hat{\mu}_c^i, \mu^*\right) + \sum_{k \neq i} \frac{t_c^k}{t_c^i } d^+\left(\hat{\mu}_c^k, \mu^* - \beta_c\right)] \leq \log n + a \log \log n \right\}\times X_n ]
\]
\[\leq \sum_{n=1}^T [\mathbb{I}\left\{t_c^i[ d\left(\hat{\mu}_c^i, \mu^*\right) + \sum_{k \neq i} \frac{\alpha_c^i}{\alpha_c^k } d^+\left(\hat{\mu}_c^k, \mu^* - \beta_c\right)] \leq \log n + a \log \log n + O(\epsilon) \right\}\times X_n] 
\]
\[\leq \sum_{n=1}^T \sum_{s=1}^n [\mathbb{I}\left\{s[ d\left(\hat{\mu}_c^i, \mu^*\right) + \sum_{k \neq i} \frac{\alpha_c^i}{\alpha_c^k } d^+\left(\hat{\mu}_c^k, \mu^* - \beta_c\right)] \leq \log T + a \log \log T + O(\epsilon) \right\}\times Y_n]
\]
where
\[
Y_n = \mathbb{I}\left\{A_n = (i,c), t_c^i(n) = s\right\}
\]
\[
 = \sum_{s=1}^T  \mathbb{I}\left\{s[ d\left(\hat{\mu}_c^i, \mu^*\right) + \sum_{k \neq i} \frac{\alpha_c^i}{\alpha_c^k } d^+\left(\hat{\mu}_c^k, \mu^* - \beta_c\right)] \leq \log T + a \log \log T + O(\epsilon) \right\}\sum_{n=s}^TY_n 
\]
\[\leq \sum_{s=1}^{\infty} \mathbb{I}\Bigg\{ s d\left(\hat{\mu}_c^i, \mu^*\right) + \sum_{k \neq i} \frac{s \alpha_c^i}{\alpha_c^k} d^+\left(\hat{\mu}_c^k, \mu^* - \beta_c\right) \leq \log T + a \log \log T  + O(\epsilon)\Bigg\} 
\]
\[\leq \lambda_c^i +\sum_{s = \lambda_c^i + 1}^{\infty} \mathbb{I}\left\{ \lambda_c^i d\left(\hat{\mu}_c^i, \mu^*\right) + \sum_{k \neq i}\frac{\lambda_c^i \alpha_c^i}{\alpha_c^k} d^+\left(\hat{\mu}_c^k, \mu^* - \beta_c\right) \leq \log T +a \log \log T + O(\epsilon) \right\},
\]
where $\lambda_c^i = \frac{\log T + a \log \log T }{\alpha_c^i L}(1 + \epsilon)$.

\[
\begin{aligned}
\alpha_c^i &= d\left(\mu_c^i, \mu^*\right) - d^+\left(\mu_c^i, \mu^* - \beta_c\right) \\
L_c &= 1 + \sum_{i=1}^{k_c} \frac{b_c^i}{\alpha_c^i}, \quad b_c^i = d^+\left(\mu_c^k, \mu^* - \beta_c\right)
\end{aligned}
\]

Therefore, the expectation bound becomes
\[
\lambda_c^i + \sum_{s = \lambda_c^i + 1}^{\infty} \mathbb{P} \left[ d\left(\hat{\mu}_c^i, \mu^*\right) + \sum_{k \neq i} \frac{\alpha_c^i}{\alpha_c^k} d^+\left(\hat{\mu}_c^k, \mu^*-\beta_c\right) \leq \frac{d^+\left(\mu_c^i, \mu^*\right) + \sum \frac{\alpha_c^i}{\alpha_c^k} d^+\left(\mu_c^k, \mu^*-\beta_c\right)}{1+\epsilon '} \right].
\]
Now, 
\[
     \sum_{s = \lambda_c^i + 1}^{\infty} \mathbb{P} \left[ d\left(\hat{\mu}_c^i, \mu^*\right) + \sum_{k \neq i} \frac{\alpha_c^i}{\alpha_c^k} d^+\left(\hat{\mu}_c^k, \mu^*-\beta_c\right) \leq \frac{d^+\left(\mu_c^i, \mu^*\right) + \sum \frac{\alpha_c^i}{\alpha_c^k} d^+\left(\mu_c^k, \mu^*-\beta_c\right)}{1+\epsilon'} \right]
\]
\[
  \leq \sum_{s = \lambda_c^i + 1}^{\infty} \mathbb{P} \left[ d\left(\hat{\mu}_c^i, \mu^*\right) \leq \frac{d\left(\hat{\mu}_c^i, \mu^*\right)}{1+\epsilon}\cup \bigcup_{k\neq i}  d^+\left(\hat{\mu}_c^k, \mu^*-\beta_c\right) \leq \frac{d^+\left(\mu_c^k, \mu^*-\beta_c\right)}{1+\epsilon'} \right].
\]

We can union bound this and use Lemma 8 from \citet{garivier2011kl}. Thus, the inequality continues as
\[
\leq \lambda_c^i + O(\epsilon').
\]

Also, since all the Clus-UCBs converge to $\mu^*$ and all the empirical means converge to their actual means, we have $f_i(n) = f_k(n)$ as n tends to infinity. Thus, for any other arm in the cluster, 
\[
\lim_{T \rightarrow \infty}\frac{\mathbb{E}[t_c^k(T)]}{\log T} = \frac{\alpha_c^i}{\alpha_c^k}\frac{1}{\alpha_c^i L_c} = \frac{1}{\alpha_c^k L_c}.
\]
\end{document}